\documentclass[letterpaper, 10 pt, conference]{ieeeconf}  
\usepackage[spanish]{babel}
\usepackage[utf8]{inputenc}
\usepackage{listings}
\usepackage{caption}
\usepackage{DejaVuSans}
\usepackage{DejaVuSansMono}
\usepackage{amssymb}
\usepackage{amsmath}

\usepackage{subcaption}
\usepackage{graphicx}

\usepackage{alltt}

\usepackage{fancyvrb}

\usepackage[table,xcdraw]{xcolor}
\usepackage{multirow}

\usepackage[T1]{fontenc}
\usepackage{fancyvrb}
\usepackage{listings}
\usepackage[scaled=.55]{beramono}    
\fvset{baselinestretch=0.5}
\DeclareCaptionFormat{listing}{\rule{\dimexpr0.9\columnwidth+17pt\relax}{0.35pt}\par\vskip1pt#1#2#3}
\newsavebox{\FVerbBox}
\usepackage[most]{tcolorbox}
\tcbset{
    frame code={}
    center title,
    left=0pt,
    right=0pt,
    top=0pt,
    bottom=0pt,
    colback=gray!10,
    colframe=white,
    width=0.45\textwidth,
    enlarge left by=0mm,
    boxsep=5pt,
    arc=0pt,outer arc=0pt,
    }

\IEEEoverridecommandlockouts                              

\title{\LARGE \bf
Towards a distributed and real-time framework for robots: Evaluation of ROS 2.0 communications for real-time robotic applications
}
\author{\textbf{Carlos San Vicente Gutiérrez}, 
    \textbf{Lander Usategui San Juan}, \\
    \textbf{Irati Zamalloa Ugarte}, 
    \textbf{Víctor Mayoral Vilches} \\
    Erle Robotics S.L. \\
    Vitoria-Gasteiz, \\
    Álava, Spain \\
}

\usepackage{blindtext}
\usepackage{graphicx}
\usepackage{listings}
\usepackage{xcolor}

\usepackage{hyperref}
\hypersetup{
    colorlinks=true,
    linkcolor=blue,
    filecolor=magenta,      
    urlcolor=cyan,
    citecolor=blue,
}

\usepackage[normalem]{ulem}
\usepackage{lipsum}

\begin{document}
\maketitle

\begin{abstract}

In this work we present an experimental setup to show the suitability of ROS 2.0 for real-time robotic applications. We disclose an evaluation of ROS 2.0 communications in a robotic inter-component (hardware) communication case on top of Linux. We benchmark and study the worst case latencies and missed deadlines to characterize ROS 2.0 communications for real-time applications. We demonstrate experimentally how computation and network congestion impacts the communication latencies and ultimately, propose a setup that, under certain conditions, mitigates these delays and obtains bounded traffic.

\end{abstract}

\section{Introduction}
\label{introduction}


In robotic systems, tasks often need to be executed with strict timing requirements. For example, in the case of a mobile robot, if the controller has to be responsive to external events, an excessive delay may result in non-desirable consequences. Moreover, if the robot was moving with a certain speed and needs to avoid an obstacle, it must detect this and stop or correct its trajectory in a certain amount of time. Otherwise, it would likely collide and disrupt the execution of the required task. These kind of situations are rather common in robotics and must be performed within well defined timing constraints that usually require real-time capabilities\footnote{Note that there is a relevant difference between having  \emph{well defined deadlines} and having the necessity to \emph{meet such deadlines} in a strict manner, which is what real-time systems deliver.}. Such systems often have timing requirements to execute tasks or exchange data over the internal network of the robot, as it is common in the case of distributed systems. This is, for example, the case of the Robot Operating System (ROS)\cite{quigley2009ros}. Not meeting the timing requirements implies that, either the system's behavior will degrade, or the system will lead to failure. 

Real-time systems can be classified depending on how critical to meet the corresponding timing constraints. For hard real-time systems, missing a deadline is considered a system failure. Examples of real-time systems are anti-lock brakes or aircraft control systems. On the other hand, firm real-time systems are more relaxed. An information or computation delivered after a missing a deadline is considered invalid, but it does not necessarily lead to system failure. In this case, missing deadlines could degrade the performance of the system. In other words, the system can tolerate a certain amount of missed deadlines before failing. Examples of firm real-time systems include 
most professional and industrial robot control systems such as the control loops of collaborative robot arms, aerial robot autopilots or most mobile robots, including self-driving vehicles. 

Finally, in the case of soft real-time, missed deadlines -even if delivered late- remain useful. This implies that soft real-time systems do not necessarily fail due to missed deadlines, instead, they produce a degradation in the usefulness of the real-time task in execution. Examples of soft-real time systems are telepresence robots of any kind (audio, video, etc.).\\ 

As ROS became the standard software infrastructure for the development of robotic applications, there was an increasing demand in the ROS community to include real-time capabilities in the framework. As a response, ROS 2.0 was created to be able to deliver real-time performance, however, as covered in previous work \cite{DBLP:journals/corr/abs-1804-07643} and \cite{2018arXiv180810821G}, the ROS 2.0 itself needs to be surrounded with the appropriate elements to deliver a complete distributed and real-time solution for robots.\\

For distributed real-time systems, communications need to provide Quality of Services (QoS) capabilities in order to guarantee deterministic end-to-end communications. ROS 2 communications use Data Distribution Service (DDS) as its communication middleware. DDS contains configurable QoS parameters which can be tuned for real-time applications. Commonly, DDS distributions use the Real Time Publish Subscribe protocol (RTPS) as a transport protocol which encapsulates the well known User Datagram Protocol (UDP). In Linux based systems, DDS implementations typically use the Linux Networking Stack (LNS) for communications over Ethernet.\\  

In previous work \cite{DBLP:journals/corr/abs-1804-07643}, we analyzed the use of layer 2 \emph{Quality of Service (QoS)} techniques such as package prioritization and Time Sensitive Networking (TSN) scheduling mechanisms to bound end-to-end latencies in Ethernet switched networks. In \cite{2018arXiv180810821G}, we analyzed the real-time performance of the LNS in a Linux PREEMPT-RT kernel and observed some of the current limitations for deterministic communications over the LNS in mixed-critical traffic scenarios. The next logical step was to analyze the real-time performance of ROS 2.0 communications in a PREEEMPT-RT kernel over Ethernet. Previous work \cite{7743223} which investigated the performance of ROS 2.0 communication showed promising results and discussed future improvements. However, the mentioned study does not explore the suitability of ROS 2.0 for real-time applications and the evaluation was not performed on an embedded platform.\\

In this work, we focus on the evaluation of ROS 2.0 communications in a robotic inter-component communication use-case. For this purpose, we are going to present a setup and a set of benchmarks where we will measure the end-to-end latencies of two ROS 2.0 nodes running in different static load conditions. We will focus our attention on worst case latencies and missed deadlines to observe the suitability of ROS 2.0 communications for real-time applications. We will also try to show the impact of different stressing conditions in ROS 2.0 traffic latencies. Ultimately, we attempt to find a suitable configuration to improve determinism of ROS 2.0 and establish the limits for such setup in an embedded platform.\\

The content below is structured as follows: section \ref{background} presents some background of ROS 2.0 and how its underlying communication middleware is structured. Section \ref{setup_results} shows the experimental results obtained while using four different approaches. Finally, Section \ref{conclusions} provides a discussion of the results.\\

\section{Background} 
\label{background}


\begin{figure}[h!]
\centering
 \includegraphics[width=0.4\textwidth]{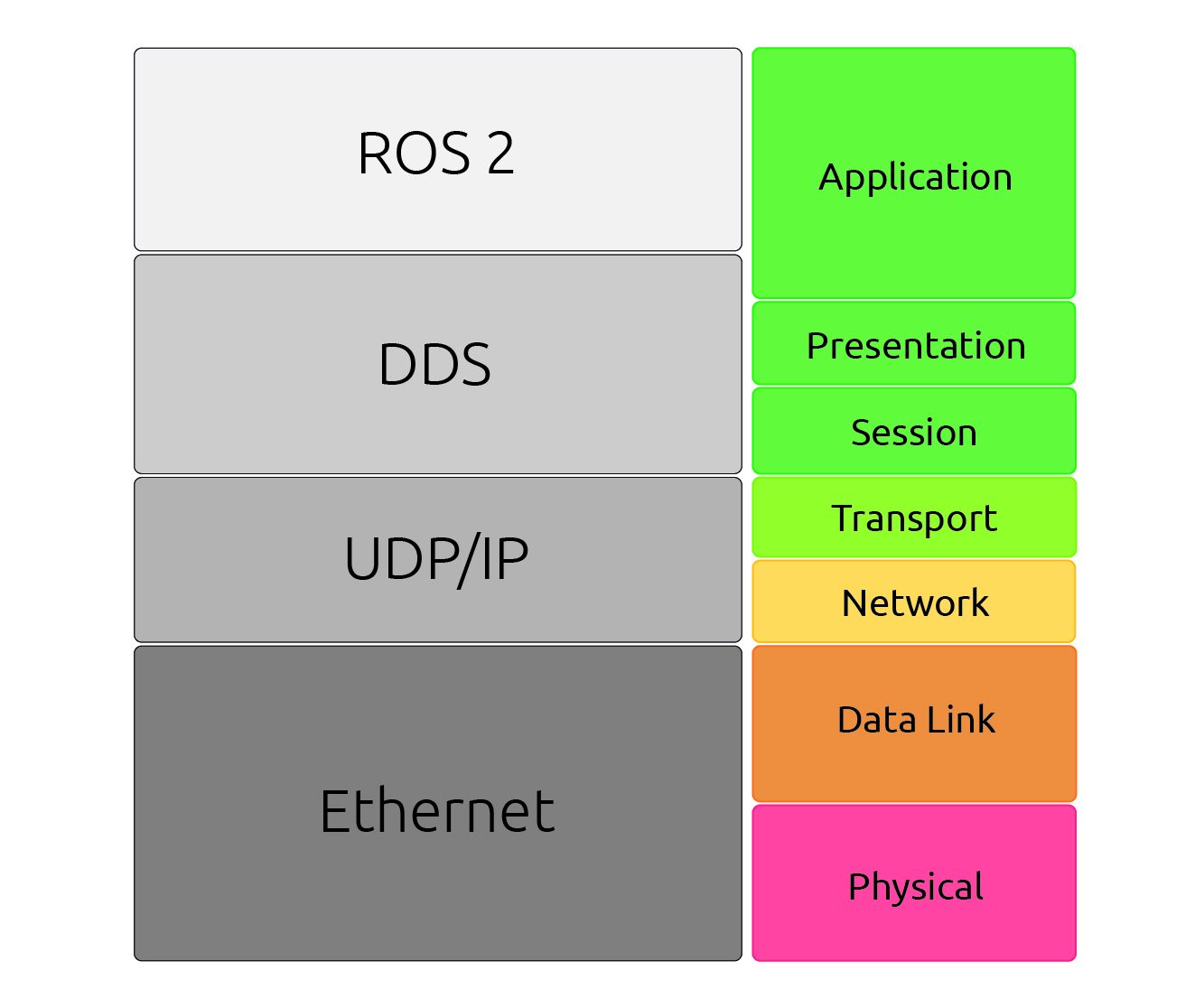}
\caption{\footnotesize Overview of ROS 2 stack for machine to machine communications over Ethernet}
\label{ros2_stack}
\end{figure}

ROS is a framework for the development of robot applications. A toolbox filled with utilities for robots, such as a communication infrastructure including standard message definitions, drivers for a variety of software and hardware components, libraries for diagnostics, navigation, manipulation and many more. Altogether, ROS simplifies the task of creating complex and robust robot behavior across a wide variety of robotic platforms. ROS 2.0 is the new version of ROS which extends the initial concept (originally meant for purely research purposes) and aims to provide a distributed and modular solution for situations involving teams of robots, real-time systems or production environments, amidst others.


Among the technical novelties introduced in ROS 2.0, Open Robotics explored several options for the ROS 2.0 communication system. They decided to use the DDS middleware due to its characteristics and benefits compared to other solutions. As documented in \cite{ros2design}, the benefit of using an end-to-end middleware, such as DDS, is that there is less code to maintain. DDS is used as a communications middleware in ROS 2.0 and it typically runs as userspace code. Even though DDS has specified standards, third parties can review audit, and implement the middleware with varying degrees of interoperability.\\

As pointed out in the technical report \cite{osrf_rt}, to have real-time performance, both a deterministic user code and an real-time operating system are needed. In our case, we will use a PREEMPT-RT patched Linux kernel as the core of our operating system for the experiments. Following the programming guidelines of the PREEMPT-RT and with a suitable kernel configuration, other authors\cite{Cerqueira_acomparison} demonstrated that it is possible to achieve system latency responses between 10 and 100 microseconds. \\

Normally, by default, DDS implementations use the Linux Network Stack (LNS) as transport and network layer. This makes the LNS a critical part for ROS 2.0 performance. However, the network stack is not optimized for bounded latencies but instead, for throughput at a given moment. In other words, there will be some limitations due to the current status of the networking stack. Nevertheless, LNS provides QoS mechanisms and thread tuning which allows to improve the determinism of critical traffic at the kernel level. \\

An important part of how the packets are processed in the Linux kernel relates actually to how hardware interrupts are handled. In a normal Linux kernel, hardware interrupts are served in two phases. In the first, an Interrupt Service Routine (ISR) is invoked when an interrupt fires, then, the hardware interrupt gets acknowledged and the work is postponed to be executed later. In a second phase, the soft interrupt, or ``bottom half'' is executed later to process the data coming from the hardware device. In PREEMPT-RT kernels, most ISRs are forced to run in threads specifically created for the interrupt. These threads are called IRQ threads \cite{lwn_irq_threads}. By handling IRQs as kernel threads, PREEMPT-RT kernels allow to schedule IRQs as user tasks, setting the priority and CPU affinity to be managed individually. IRQ handlers running in threads can themselves be interrupted so the latency due to interrupts is mitigated. For our particular interests, since our application needs to send critical traffic, it is possible to set the priority of the Ethernet interrupt threads higher than other IRQ threads to improve the network determinism.\\

Another important difference between a normal and a PREEMPT-RT kernel is within the context where the softirq are executed. Starting from kernel version 3.6.1-rt1 on, the soft IRQ handlers are executed in the context of the thread that raised that Soft IRQ \cite{lwn_softirq}. Consequently, the NET\_RX soft IRQ, which is the softirq for receiving network packets, will normally be executed in the context of the network device IRQ thread. This allows a fine control of the networking processing context. However, if the network IRQ thread is preempted or it exhausts its NAPI\footnote{`New API' or NAPI for short is an extension to the device driver packet processing framework, which is designed to improve the performance of high-speed networking.} weight time slice, it is executed in the ksoftirqd/n (where n is the logical number of the CPU). \\

Processing packets in ksoftirqd/n context is troublesome for real-time because this thread is used by different processes for deferred work and can add latency. Also, as the ksoftirqd thread runs with SCHED\_OTHER policy, it can be easily preempted. In practice, the soft IRQs are normally executed in the context of the Ethernet IRQ threads and in the ksoftirqd/n thread, for high network loads and under heavy stress (CPU, memory, I/O, etc.). The conclusion here is that, in normal conditions, we can expect reasonable deterministic behavior, but if the network and the system are loaded, the latencies can increase greatly. \\  


\section{Experimental setup and results}
\label{setup_results}

This section presents the setup used to evaluate the real-time performance of ROS 2.0 communications over Ethernet in a PREEMPT-RT patched kernel. Specifically, we measure the end-to-end latencies between two ROS 2.0 nodes in different machines. For the experimental setup, we used a GNU/Linux PC and an embedded device which could represent a robot controller (RC) and a robot component (C) respectively. \\

The PC used for the tests has the following characteristics:
\begin{itemize}
    \item Processor: Intel(R) Core(TM) i7-8700K CPU @ 3.70GHz (6 cores).
    \item OS: Ubuntu 16.04 (Xenial).
    \item ROS 2.0 version: ardent.
    \item Kernel version: 4.9.30.
    \item PREEMPT-RT patch: rt21.
    \item Link capacity: 100/1000 Mbps, Full-Duplex.
    \item NIC: Intel i210.
\end{itemize}

In the other hand, the main characteristics of the embedded device are:
\begin{itemize}
    \item Processor: ARMv7 Processor (2 cores).
    \item ROS 2 version: ardent 
    \item Kernel version: 4.9.30.
    \item PREEMPT-RT patch: rt21.
    \item Link capacity: 100/1000 Mbps, Full-Duplex.
\end{itemize}

\begin{figure}[h!]
\centering
 \includegraphics[width=0.5
\textwidth]{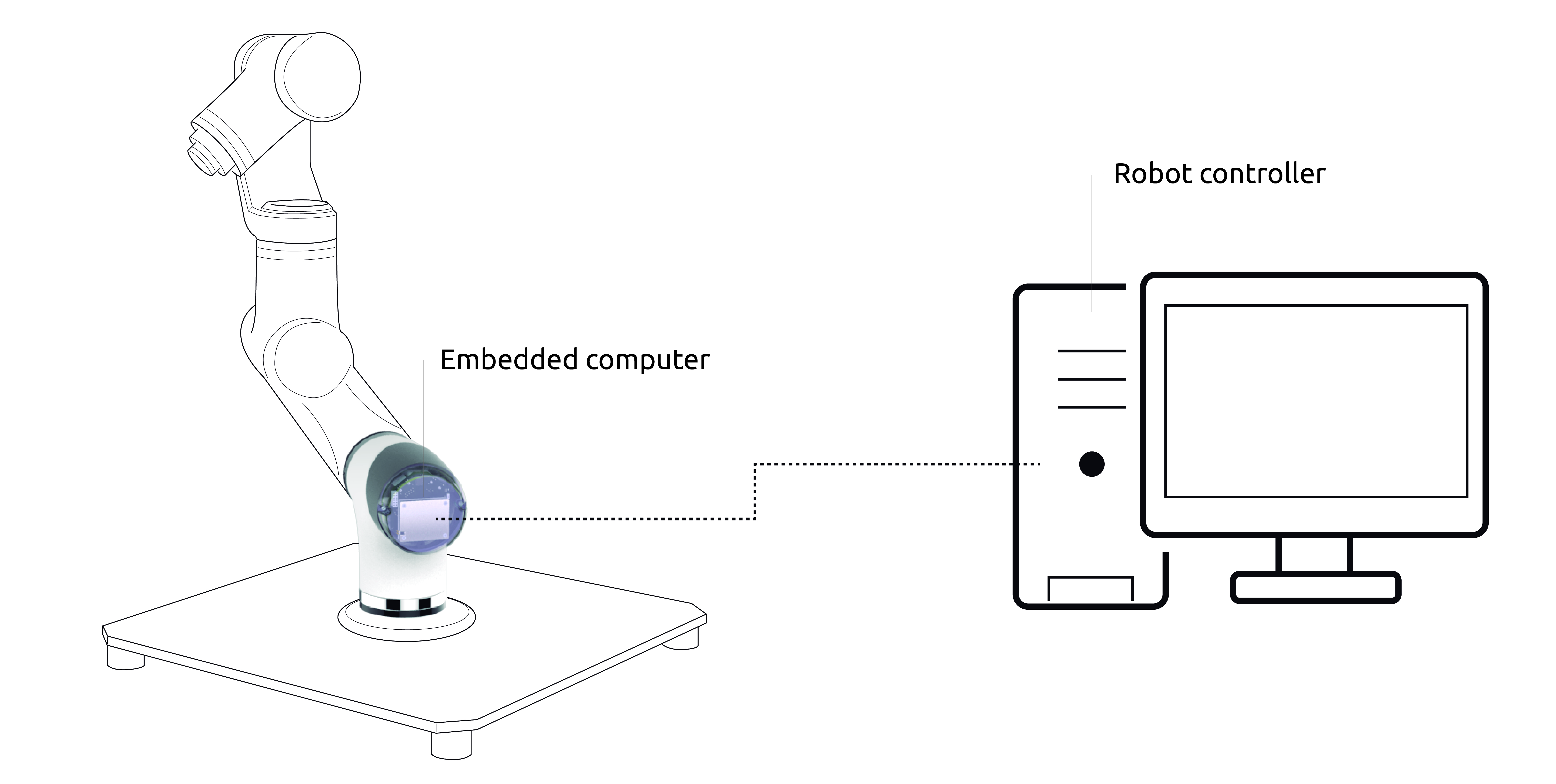}
\caption{\footnotesize Experimental setup overview.}
\label{scenario}
\end{figure}

The RC and the C are connected point to point using a CAT6e Ethernet wire as shown in figure \ref{scenario}. 



\subsection{Round-trip test setup}
\label{round-trip}

\begin{figure}[h!]
\centering
 \includegraphics[width=0.4\textwidth]{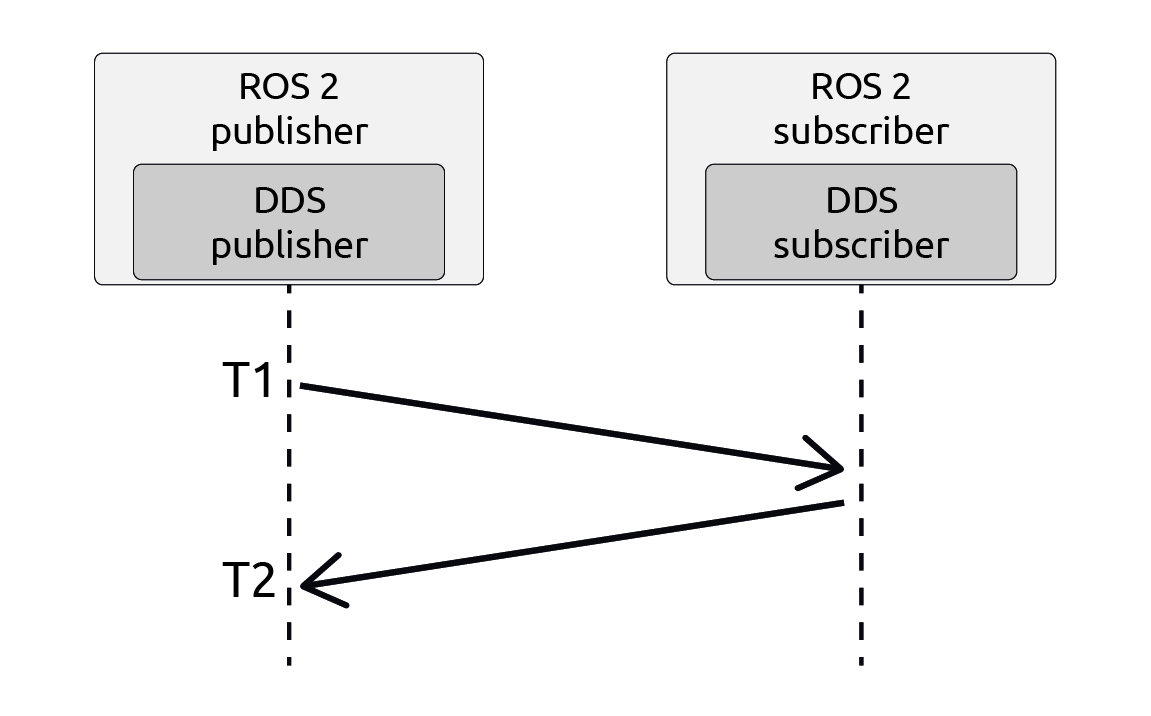}
\caption{\footnotesize Graphical presentation of the measured DDS round-trip latency. T1 is the time-stamp when data is send from the DDS publisher and T 2 is the time-stamp when data is received at the DDS publisher. Round-trip latency is defined as T2 - T1.}
\label{ping-pong}
\end{figure}

The communications between the robot controller and the robot component are evaluated with a round-trip time (RTT) test, also called ping-pong test. We use a ROS 2.0 node as the client in RC and a ROS 2.0 node as the server in C. The round-trip latency is measured as the time it takes for a message to travel from the client to the server, and from the server back to the client. The message latency is measured as the difference between the time-stamp taken before sending the message (T1) in the client and the time-stamp taken just after the reception of the message in the callback of client (T2), as shown in figure \ref{ping-pong}. \\

The client creates a publisher that is used to send a message through a topic named `ping'. The client also creates a subscriber, which waits for a reply in a topic named `pong'. The server creates a subscriber which waits for the `ping' topic and a publisher that replies the same received message through the `pong' topic. The `ping' and `pong' topics are described by a custom message with an UINT64 used to write the message sequence number and a U8[] array to create a variable payload.\\ 

\begin{figure}[h!]
\centering
 \includegraphics[width=0.5\textwidth]{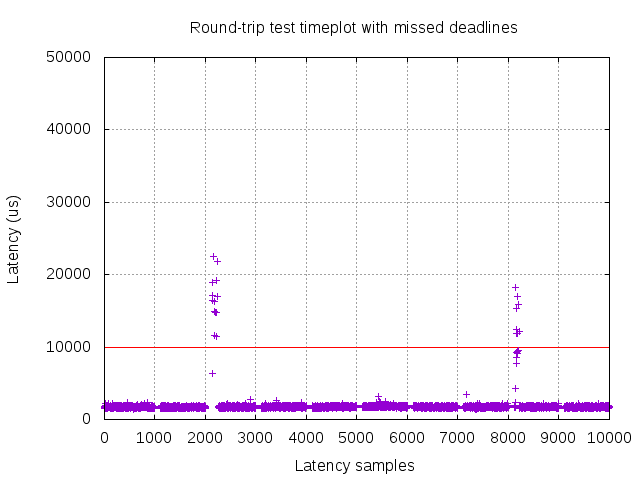}
\caption{\footnotesize Time plot example with a 10 millisecond deadline and missed deadlines. (The image corresponds to Test2.D with DDS2)}
\label{overrun}
\end{figure}

For the tests performed in this work, we used a publishing rate of 10 milliseconds. The sending time is configured by waking the publisher thread with an absolute time. If the reply arrives later than 10 milliseconds and the current sending time has expired, the message is published in the next available cycle. If there is no reply in 500 milliseconds, we consider that the message was lost. Also, if the measured round-trip latency is higher than 10 milliseconds, we consider it a missed deadline, as shown in figure \ref{overrun}. \\

For all the experiments we have used the same ROS 2.0 QoS profile. We followed the guidelines described in the inverted pendulum control of the ROS 2.0 demos \cite{pendulum_demo}: best-effort reliability, KEEP\_LAST history and a history depth of 1. This configuration is not optimized for reliability but for low latencies. One of the motivations behind this configuration is that the reliable mode can potentially block, which is an unwanted behavior in real-time threads.\\

All the experiments are run using three different DDS implementations.\\

\subsection{Experimental results}
\label{results}


\subsubsection{Test 1. System under load}
\label{exp1}
In this test, we want to explore how communications get affected when the system is under heavy load. Additionally, we aim to show how a proper real-time configuration can mitigate these effects.\\

For the `ping' and `pong' topics messages, we use a payload of 500 Bytes \footnote{This makes a total packet size of 630 Bytes, summing the sequence number sub-message and UDP and RTPS headers.}. To generate load in the system we use the tool `\emph{stress}' generating CPU stress, with 8 CPU workers, 8 VM workers, 8 I/O workers, and 8 HDD workers in the PC and 2 workers per stress type in the embedded device.\\

We run the following tests:\\

\begin{itemize}
    \item Test1.A: System idle with no real-time settings. 
    \item Test1.B: System under load with no real-time settings. 
    \item Test1.C: System idle with real-time settings. 
    \item Test1.D: System under load without real-time settings.  
\end{itemize}

\begin{figure*}[h!]
  \begin{subfigure}[t]{.5\textwidth}
    \centering
    \includegraphics[width=0.8\linewidth]{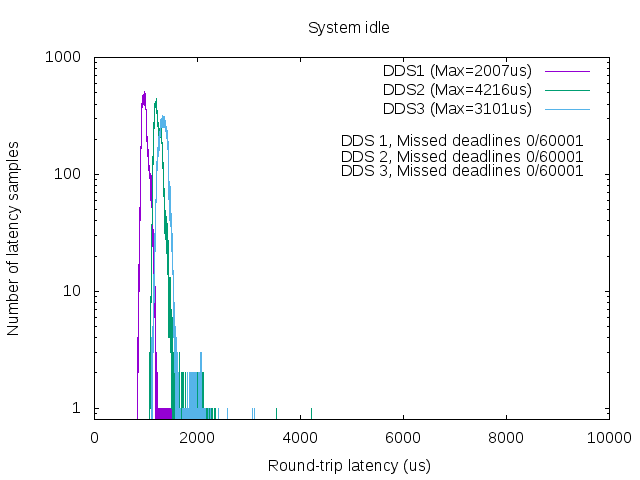}
    \caption{System idle.}
   \label{Test1.A}
  \end{subfigure}
  \hfill
  \begin{subfigure}[t]{.5\textwidth}
    \centering
    \includegraphics[width=0.8\linewidth]{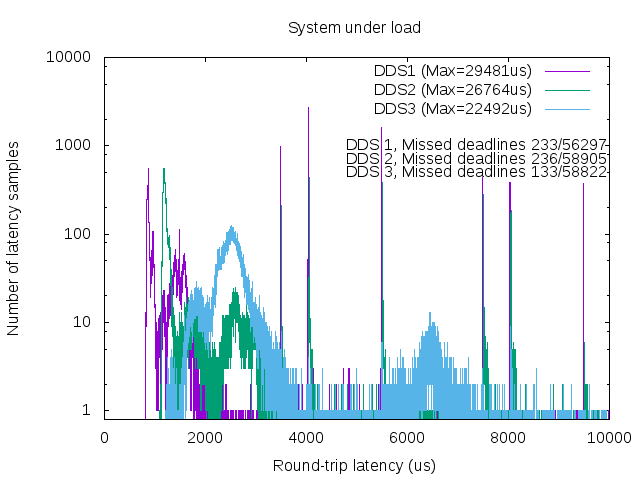}
    \caption{System under load.}
    \label{Test1.B}
  \end{subfigure}
  
   \medskip

  \begin{subfigure}[t]{.5\textwidth}
    \centering
    \includegraphics[width=0.8\linewidth]{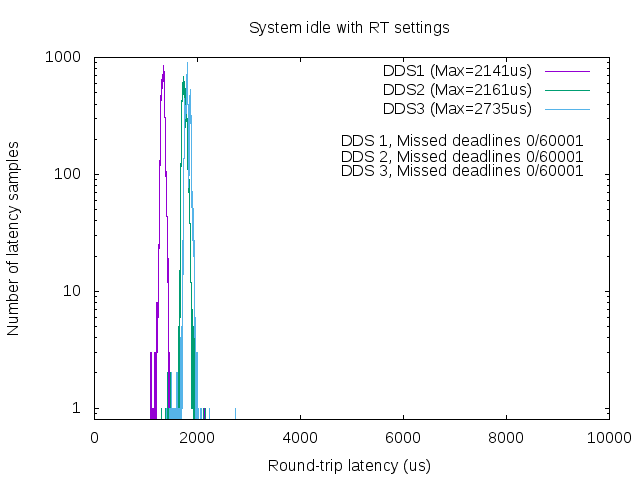}
    \caption{System idle with RT settings.}
    \label{Test1.C}
  \end{subfigure}
  \hfill
  \begin{subfigure}[t]{.5\textwidth}
    \centering
    \includegraphics[width=0.8\linewidth]{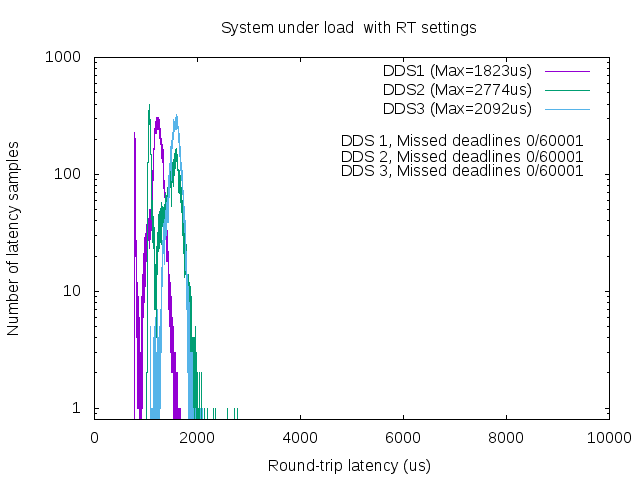}
    \caption{System under load with RT settings.}
    \label{Test1.D}
  \end{subfigure} 
  \caption{\footnotesize Impact of RT settings under different system load. a) System without additional load without RT settings. b) System under load without RT settings. c) System without additional load and RT settings. d) System under load and RT settings.}
\end{figure*}

In the experiment Test1.A we run the round-trip test in normal conditions (Idle), that is, no other processes apart from the system default are running during the test. Figure \ref{Test1.A} shows stable latency values with a reasonable low jitter (table \ref{tab:idle}). We get latencies higher than 4 milliseconds for an specific DDS, which means that we probably can expect higher worst case latencies with a longer test duration. \\ 

In the experiment Test1.B, we run the round-trip test with the system under load. Figure \ref{Test1.B} shows how, in this case, latencies are severely affected resulting in a high number of missed deadlines (table \ref{tab:stress}). This happens mainly because all the program threads are running with SCHED\_OTHER scheduling and are contending with the stress program for CPU time. Also, some latencies might also be caused due to memory page faults causing the memory not to be locked. The results show how sensitive non-real time processes are to system loads. As we wait until the arrival of the messages to send the next and we are fixing a 10 minutes duration for the test, the number of messages sent during each test might be different if deadlines are missed.  \\ 

In the experiment Test1.C we repeat the round-trip test configuring the application with real-time properties. The configuration used includes memory locking, the use of the TSLF allocators for the ROS 2.0 publishers and subscriptors,  as well as running the round-trip program threads with SCHED\_FIFO scheduling and real-time priorities. The configuration of the DDS threads has been done in a different manner, depending on each DDS implementation. Some DDS implementations allowed to configure their thread priorities using the Quality of Service (QoS) profiles (through XML files), while others did not. In the cases where it was feasible, it allowed a finer real-time tuning since we could set higher priorities for the most relevant threads, as well as configuring other relevant features such memory locking and buffer pre-allocation. For the DDS implementations in which it was not possible to set the DDS thread priorities, we set a real-time priority for the main thread, so that all the threads created did inherit this priority.\\

Figure \ref{Test1.D} shows a clear improvement for all the DDS implementations when comparing to case Test1.B (table \ref{tab:stress_rt}), where we did not use any real-time settings. Henceforth, all the tests in the following sections are run using the real-time configurations.

\subsubsection{Test 2. Concurrent traffic}
\label{exp2}

%

In this second test, we want to observe the impact of non-critical traffic in the round-trip test measurements. We generate certain amount of traffic from RC to C and from C to RC using the tool \emph{iperf}. \footnote{When the system was stressed the embedded devices were not capable of generating more than 20 Mbps steadily.}\\

In this experiment, we do not prioritize the traffic, which means the concurrent traffic will use the same Qdisc and interrupt threads than the critical traffic. In other words, we mix both non-critical and critical traffic together and analyze the impact of simply doing so.\\ 

We run the test generating 1 Mbps, 40 Mbps and 80 Mbps concurrent traffic with and without stressing the system:

\begin{itemize}
    \item Test2.A: System idle with 1 Mbps concurrent traffic. 
    \item Test2.B: System under load with 1 Mbps concurrent traffic. 
    \item Test2.C: System idle with 40 Mbps concurrent traffic. 
    \item Test2.D: System under load with 40 Mbps concurrent traffic. 
    \item Test2.E: System idle with 80 Mbps concurrent traffic. 
    \item Test2.F: System under load with 80 Mbps concurrent traffic.     
\end{itemize}

\begin{figure*}[h!]
 \begin{subfigure}[t]{.5\textwidth}
    \centering
    \includegraphics[width=0.8\linewidth]{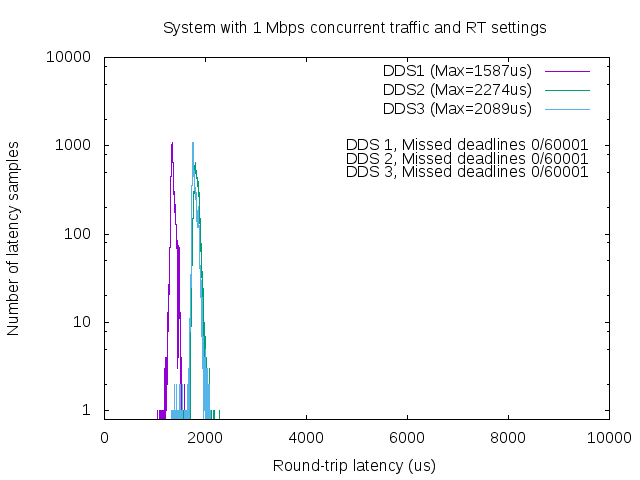}
    \caption{System with 1 Mbps concurrent traffic and RT settings.}
    \label{Test2.A}
  \end{subfigure}
  \hfill
  \begin{subfigure}[t]{.5\textwidth}
    \centering
    \includegraphics[width=0.8\linewidth]{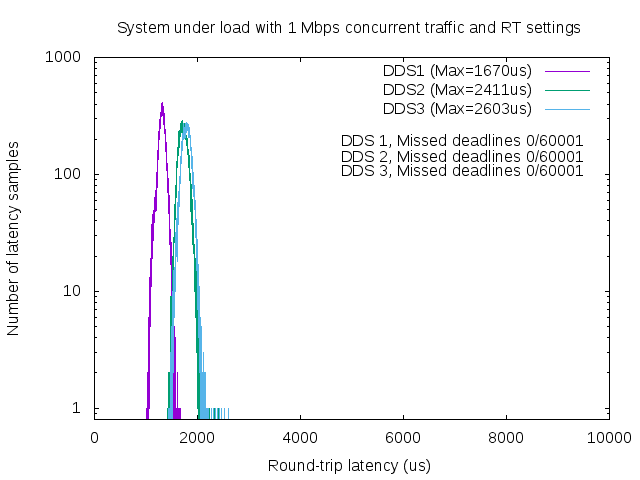}
    \caption{System under load with 1 Mbps concurrent traffic and RT settings.}
    \label{Test2.B}
  \end{subfigure}
  
   \medskip

  \begin{subfigure}[t]{.5\textwidth}
    \centering
    \includegraphics[width=0.8\linewidth]{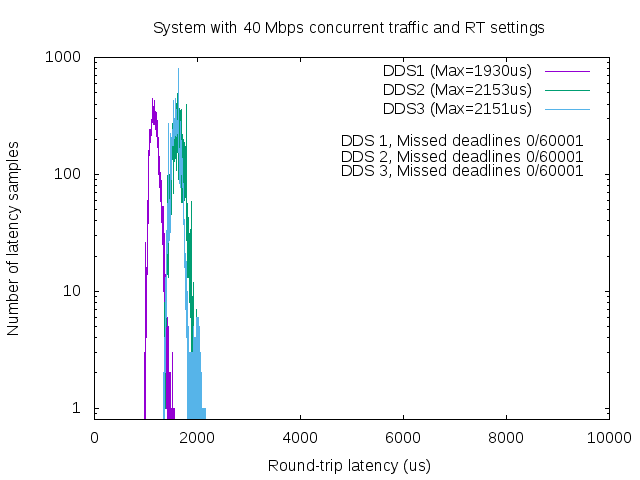}
    \caption{System with 40 Mbps concurrent traffic and RT settings.}
    \label{Test2.C}
  \end{subfigure}
  \hfill
  \begin{subfigure}[t]{.5\textwidth}
    \centering
    \includegraphics[width=0.8\linewidth]{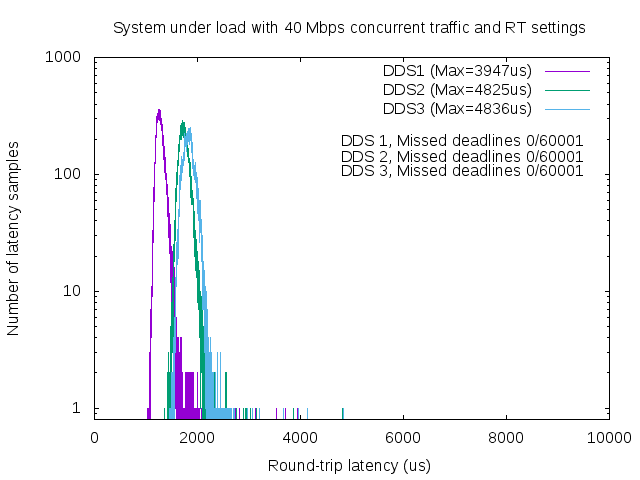}
    \caption{System under load with 40 Mbps concurrent traffic and RT settings.}
    \label{Test2.D}
  \end{subfigure}
  
   \medskip

  \begin{subfigure}[t]{.5\textwidth}
    \centering
    \includegraphics[width=0.8\linewidth]{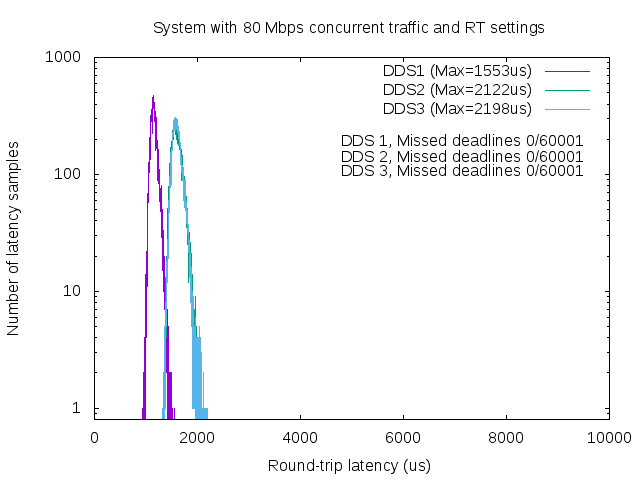}
    \caption{System with 80 Mbps concurrent traffic and RT settings.}
    \label{Test2.E}
  \end{subfigure}
  \hfill
  \begin{subfigure}[t]{.5\textwidth}
    \centering
    \includegraphics[width=0.8\linewidth]{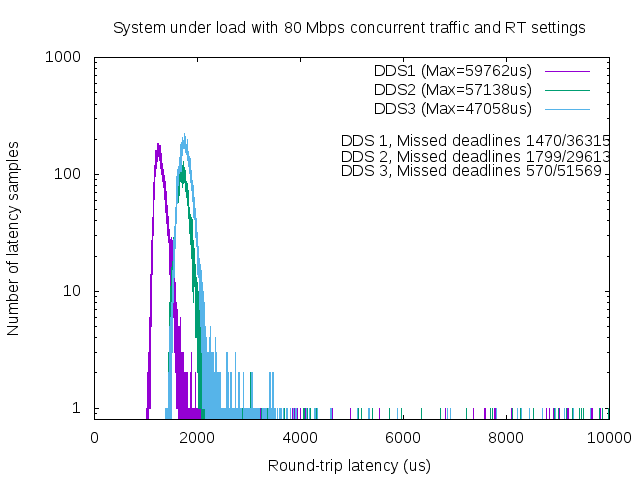}
    \caption{System under load with 80 Mbps concurrent traffic and RT settings.}
    \label{Test2.F}
  \end{subfigure} 
  \caption{\footnotesize Impact of concurrent traffic with RT settings. a) System without additional load and 40 Mbps of concurrent traffic. b) System under load and 40 Mbps of concurrent traffic. c) System without additional load and 80 Mbps of concurrent traffic. d) System under load and 80 Mbps of concurrent traffic.}
\end{figure*}

For the non stressed cases Test2.A, Test2.C and Test2.E, we observe that the concurrent traffic does not affect the test latencies significantly (figures \ref{Test2.A}, \ref{Test2.C}, \ref{Test2.E}). When we stress the system, we can see how for 1 Mbps and 40 Mbps the latencies are still under 10 milliseconds. However, for 80 Mbps, the communications are highly affected resulting in a high number of lost messages and missed deadlines.\\ 

As we explained in \ref{background}, depending on the network load, packets might be processed in the \emph{ksoftirqd} threads. Because these threads run with no real-time priority, they would be highly affected by the system load. For this reason, we can expect higher latencies when both, concurrent traffic and system loads, are combined. For high network loads, the context switch would occur more frequently or even permanently. For medium network loads, the context switch happens intermittently with a frequency that correlates directly with the exact network load circumstances at each given time.\\

\subsubsection{Test 3. Increasing message payload size}
\label{exp3}

In this third experiment, we increase the `Ping' topic message payload to observe the determinism for higher critical bandwidth traffic. By increasing the payload, messages get fragmented. This implies that higher bandwidth is used and, as it happened in the previous test\ref{exp2}, depending on the traffic load, the packets may be processed in the \emph{ksoftirqd} threads or not. Once again, we expect this action to have negative consequences to bound latencies.\\ 

For this test we use two different payload sizes: 32 Kbytes and 128 Kbytes.

\begin{itemize}
    \item Test3.A: System idle with 32 Kbytes payload. 
    \item Test3.B: System under load with 32 Kbytes payload. 
    \item Test3.C: System under load with 128 Kbytes payload. 
\end{itemize}

\begin{figure*}[h!]
  \begin{subfigure}[t]{.5\textwidth}
    \centering
    \includegraphics[width=0.8\linewidth]{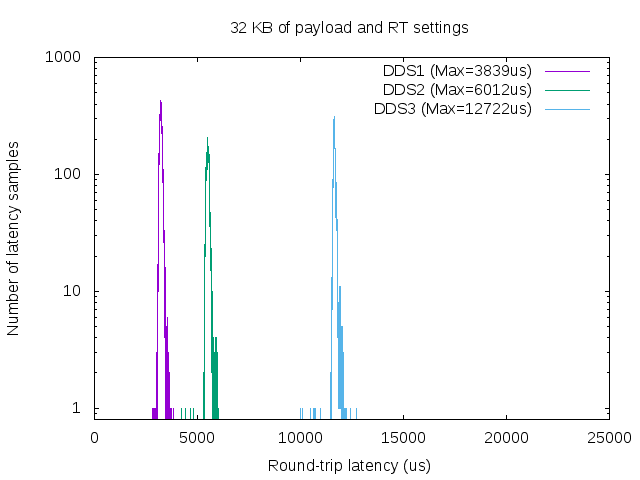}
    \caption{32 KB of payload with RT settings.}
    \label{Test3.A}
  \end{subfigure}
  \hfill
  \begin{subfigure}[t]{.5\textwidth}
    \centering
    \includegraphics[width=0.8\linewidth]{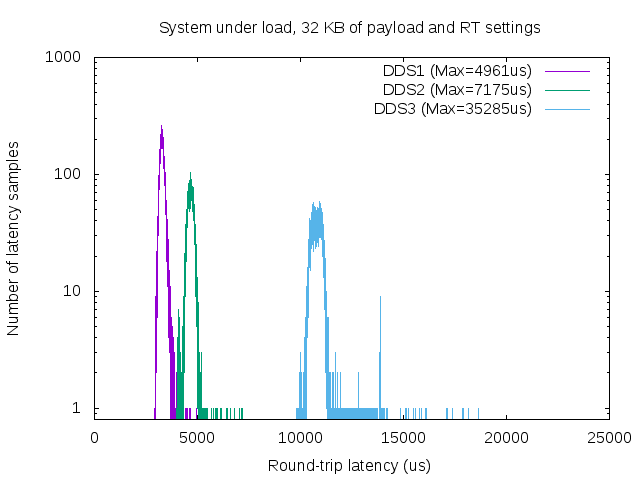}
    \caption{System under load, 32 KB of payload and RT settings.}
    \label{Test3.B}
  \end{subfigure}
  \caption{\footnotesize Impact of the different payload size under different system conditions. a) 32 KB of payload without additional system load. b) 32 KB of payload under system load.}
\end{figure*}

For Test3.A, 32 Kbytes and the system stressed, we can start observing some high latencies (figure \ref{Test3.B}), however there are no missed deadlines and message losses during a 10 minute duration test. On the other hand, for 128 Kbytes, we start observing missed deadlines and even packet loss in a 10 minutes test window. (\ref{tab:128k_stress_rt}).\\

In the previous experiment \ref{exp2}, non-critical traffic was the cause of high latencies. As the context switch from the Ethernet IRQ thread to the \emph{ksoftirqd} threads occurs for a certain amount of consecutive frames, in this case the critical traffic is causing by its own the context switch. This must be taken into account when sending critical messages with high payloads.\\ 

It is worth noting, that for this experiment, the DDS QoS were not optimized for throughput. Also, the publishing rates could not be maintained in a 10 millisecond window, thereby the missed deadlines statistics have been omitted in the plots.


\subsubsection{Test 4. Tuning the kernel threads priorities}
\label{exp4}

As suspected, the main problems demonstrated previously in \ref{exp2} and \ref{exp3} were caused by the packet processing (switching) in the ksoftirqd threads. In this experiment, we decided to tune the kernel threads to mitigate the problem.\\

We configured the threads with the following priorities:
\begin{itemize}
    \item Ethernet IRQ threads: priority 90
    \item ROS 2 executor threads: priority 80
    \item DDS threads: priority 70-80 \footnote{Some DDS threads were configured with specific priorities using the QoS vendor profile XML}
    \item ksoftirqd/n threads: 60
\end{itemize}

We repeated Test2.F (System under load with 80 Mbps concurrent traffic) using the new configuration.

\begin{itemize}
    \item Test4.A: System under load with 80 Mbps concurrent traffic.
\end{itemize}

\begin{figure*}[h!]
    \centering
    \includegraphics[width=0.4\linewidth]{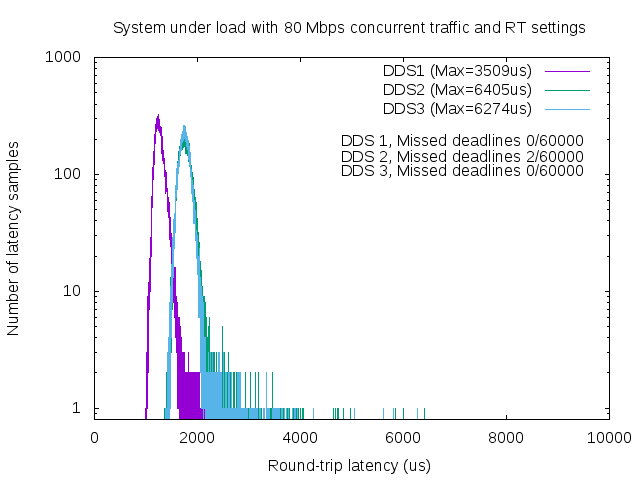}
    \caption{System under load with 80 Mbps concurrent traffic and RT settings and kernel thread tuning.}
    \label{Test4.A}
\end{figure*}

Comparing to Test2.F (Figures \ref{Test2.F} and \ref{Test4.A}) we observe a clear improvement with the new configuration. We observe almost no missed deadlines, nor message loss for a 10 minutes duration test. This suggests that the high latencies observed in \ref{exp2} and \ref{exp3} were caused because of the lack of real-time priority of \emph{ksoftirqd} when there is a high bandwidth traffic in the networking stack.\\ 


\subsubsection{Test 5. Long term test}
\label{exp5}

All the previous experiments were run within a 10 minute duration window. While this duration can be enough to evaluate the performance and identify some problems, it is usually not enough to observe the worst cases for a given configuration. Because some delays may appear just when specific events happen, not observing missed deadlines does not guarantee that it cannot happen in the long term. Therefore, we decided to repeat the test from experiment Test2.C for a 12 hour window.\\

\begin{itemize}
    \item Test5.A: System under load with 1 Mbps concurrent traffic. Duration 12 hours.
    \item Test5.B: System under load with 40 Mbps concurrent traffic. Duration 12 hours.
\end{itemize}

For 40 Mbps we observed some message lost that we did not observe in a 10 minute long test. However, the number of messages lost is reasonably low and depending on how critical is the application, it can be considered acceptable. For 1 Mbps we did not observe any message loss, nor a relevant amount of missed deadlines. We did observe a very low jitter when compared to the 40 Mbps case.

\begin{figure*}[h!]
  \begin{subfigure}[t]{.5\textwidth}
    \centering
    \includegraphics[width=0.8\linewidth]{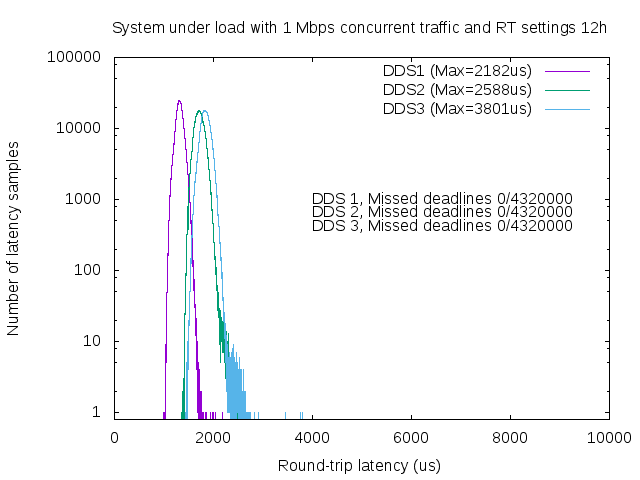}
    \caption{System under load with 1 Mbps concurrent traffic and RT settings.}
    \label{Test5.A}
  \end{subfigure}
  \hfill
  \begin{subfigure}[t]{.5\textwidth}
    \centering
    \includegraphics[width=0.8\linewidth]{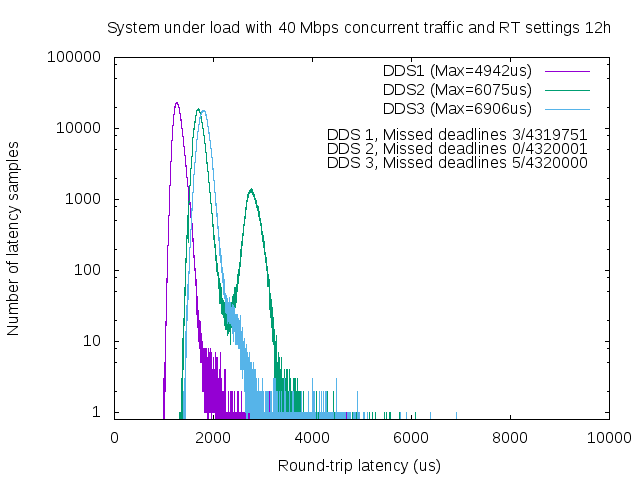}
    \caption{System under load with 40 Mbps concurrent traffic and RT settings.}
    \label{Test5.B}
  \end{subfigure}
  \caption{\footnotesize Impact of concurrent traffic with RT settings in a long term test of 12 hours duration.}
\end{figure*}

%

\section{Conclusion and future work}
\label{conclusions}

In this work, we presented an experimental setup to show the suitability of ROS 2.0 for real-time robotic applications. We have measured the end-to-end latencies of ROS 2.0 communications using different DDS middleware implementations in different stress conditions. The results showed that a proper real-time configuration of the ROS 2.0 framework and DDS threads reduces greatly the jitter and worst case latencies.\\

We also observed the limitations when there is non-critical traffic in the Linux Network Stack. Some of these problems can be avoided or minimized by configuring the network interrupt threads and using Linux traffic control QoS methods. Based on our results, we conclude that it seems possible to achieve firm and soft real-time Ethernet communications with mixed-critical traffic by using the Linux Network Stack but not hard real-time due to the observed limitations.
There is ongoing work in the Linux kernel which may eventually improve the determinism in the Linux Network Stack \cite{xdp}, \cite{AF_XDP} \cite{ETF}.\\

For the moment, the best strategies to optimize ROS 2.0 communications for real-time applications are to a) configure the application with real-time settings, b) configure the kernel threads accordingly to the application settings and c) limit the network and CPU usage of each one of the machines involved in the communication.\\

Regarding the DDS middleware implementations evaluated, we observed differences in the performance and on the average latencies. These differences may be caused by a variety of reasons such as the DDS implementation itself, but also because of the ROS 2.0 RMW layer or even by the configurations used for the experiments. Regardless of the performance, we observed a similar behavior in terms of missed deadlines and loss messages which confirms the interest of DDS for real-time scenarios. 

In future work, we will evaluate several methods to limit the network and CPU usage. One way to achieve this is using the Linux control groups (cgroups) to isolate the application in exclusive CPUs. Using cgroups, it is possible to set the priority of the application traffic using `net\_prio'. This would help to isolate critical traffic from non-critical traffic. Also, we will evaluate the impact of non-critical traffic from another ROS 2.0 node in the same process or from the same node. For that purpose, we will focus on how ROS 2.0 executors and the DDS deal with mixed-critical topics.

\newpage

\bibliographystyle{IEEEtran}
\bibliography{references}

\newpage
\onecolumn


\begin{table*}[ht]
\centering
\caption{Round-trip latency results: System idle}
\label{tab:idle}
\begin{tabular}{|c|c|c|c|c|c|}
\hline
\multicolumn{6}{|c|}{System Idle}                        \\ \hline
      & Min($\mu$s) & Avg($\mu$s) & Max($\mu$s) & Missed deadlines & Message loss \\ \hline
DDS 1 & 827    & 981    & 2007   & 0/60001       & 0/60001           \\ \hline
DDS 2 & 1059   & 1237   & 4216   & 0/60001       & 0/60001          \\ \hline
DDS 3 & 1105   & 1335   & 3101   & 0/60001       & 0/60001           \\ \hline
\end{tabular}
\end{table*}

\begin{table*}[ht]
\centering
\caption{Round-trip latency results: System idle with RT settings}
\label{tab:idle_rt}
\begin{tabular}{|c|c|c|c|c|c|}
\hline
\multicolumn{6}{|c|}{System idle with RT settings}                        \\ \hline
      & Min($\mu$s) & Avg($\mu$s) & Max($\mu$s) & Missed deadlines & Message loss \\ \hline
DDS 1 & 1079    & 1329    & 2141   & 0/60001       & 0/60001           \\ \hline
DDS 2 & 1301   & 1749   & 2161   & 0/60001       & 0/60001          \\ \hline
DDS 3 & 1394   & 1809   & 2735   & 0/60001       & 0/60001           \\ \hline
\end{tabular}
\end{table*}

\begin{table*}[ht]
\centering
\caption{Round-trip latency results: System under load}
\label{tab:stress}
\begin{tabular}{|c|c|c|c|c|c|}
\hline
\multicolumn{6}{|c|}{System under load}                        \\ \hline
      & Min($\mu$s) & Avg($\mu$s) & Max($\mu$s) & Missed deadlines & Message loss \\ \hline
DDS 1 & 810    & 2524    & 29481   & 233/56297      & 0/56297           \\ \hline
DDS 2 & 1082   & 2150   & 26764   & 236/58905       & 0/58905           \\ \hline
DDS 3 & 1214   & 2750   & 22492   & 133/58822       & 0/58822           \\ \hline
\end{tabular}
\end{table*}

\begin{table*}[ht]
\centering
\caption{Round-trip latency results: System under load with RT settings}
\label{tab:stress_rt}
\begin{tabular}{|c|c|c|c|c|c|}
\hline
\multicolumn{6}{|c|}{System under load with RT settings}                        \\ \hline
      & Min($\mu$s) & Avg($\mu$s) & Max($\mu$s) & Missed deadlines & Message loss \\ \hline
DDS 1 & 769    & 1197    & 1823   & 0/60001       & 0/60001           \\ \hline
DDS 2 & 1008   & 1378   & 2774   & 0/60001       & 0/60001           \\ \hline
DDS 3 & 1082   & 1568   & 2092   & 0/60001      & 0/60001           \\ \hline
\end{tabular}
\end{table*}

\begin{table*}[ht]
\centering
\caption{Round-trip latency results: System with 40 Mbps concurrent traffic and RT settings}
\label{tab:iperf40_rt}
\begin{tabular}{|c|c|c|c|c|c|}
\hline
\multicolumn{6}{|c|}{System with 40 Mbps concurrent traffic and RT settings} \\ \hline
      & Min($\mu$s) & Avg($\mu$s) & Max($\mu$s) & Missed deadlines & Message loss \\ \hline
DDS 1 & 976    & 1165   & 1930   & 0/60001       & 0/60001           \\ \hline
DDS 2 & 1340   & 1628   & 2153   & 0/60001       & 0/60001           \\ \hline
DDS 3 & 1343   & 1582   & 2151   & 0/60001       & 0/60001           \\ \hline
\end{tabular}
\end{table*}

\begin{table*}[ht]
\centering
\caption{Round-trip latency results: System under load with 40 Mbps concurrent traffic and RT settings}
\label{tab:iperf40_stress_rt}
\begin{tabular}{|c|c|c|c|c|c|}
\hline
\multicolumn{6}{|c|}{System under load with 40 Mbps concurrent traffic and RT settings} \\ \hline
      & Min($\mu$s) & Avg($\mu$s) & Max($\mu$s) & Missed deadlines & Message loss \\ \hline
DDS 1 & 1037    & 1283   & 3947   & 0/60001       & 0/60001           \\ \hline
DDS 2 & 1358   & 1744   & 4825   & 0/60001       & 0/60001           \\ \hline
DDS 3 & 1458   & 1835   & 4836   & 0/60001       & 0/60001           \\ \hline
\end{tabular}
\end{table*}

\begin{table*}[ht]
\centering
\caption{Round-trip latency results: System with 80 Mbps concurrent traffic and RT settings}
\label{tab:iperf80_rt}
\begin{tabular}{|c|c|c|c|c|c|}
\hline
\multicolumn{6}{|c|}{System with 80 Mbps concurrent traffic and RT settings} \\ \hline
      & Min($\mu$s) & Avg($\mu$s) & Max($\mu$s) & Missed deadlines & Message loss \\ \hline
DDS 1 & 939    & 1157   & 1553   & 0/60001       & 0/60001           \\ \hline
DDS 2 & 1333   & 1603   & 2122   & 0/60001       & 0/60001           \\ \hline
DDS 3 & 1320   & 1607   & 2198   & 0/60001       & 0/60001           \\ \hline
\end{tabular}
\end{table*}

\begin{table*}[ht]
\centering
\caption{Round-trip latency results: System under load with 80 Mbps concurrent traffic and RT settings}
\label{tab:iperf80_stress_rt}
\begin{tabular}{|c|c|c|c|c|c|}
\hline
\multicolumn{6}{|c|}{System under load with 80 Mbps concurrent traffic and RT settings} \\ \hline
      & Min($\mu$s) & Avg($\mu$s) & Max($\mu$s) & Missed deadlines & Message loss \\ \hline
DDS 1 & 1011    & 1383   & 59762   & 240/36315       & 1230/36315           \\ \hline
DDS 2 & 1375  & 1805   & 57138   & 224/29613       & 1575/29613           \\ \hline
DDS 3 & 1388   & 1803   & 47058   & 119/51569       & 451/51569           \\ \hline
\end{tabular}
\end{table*}

\begin{table*}[ht]
\centering
\caption{Round-trip latency results: System under load with 80 Mbps concurrent traffic and RT settings (ksoftirqd prio)}
\label{tab:128k_stress_rt}
\begin{tabular}{|c|c|c|c|c|c|}
\hline
\multicolumn{6}{|c|}{System under load with 80 Mbps concurrent traffic and RT settings (ksoftirqd prio)} \\ \hline
      & Min($\mu$s) & Avg($\mu$s) & Max($\mu$s) & Missed deadlines & Message loss \\ \hline
DDS 1 & 991    & 1273   & 3509   & 0/60000       & 0/60000           \\ \hline
DDS 2 & 1365  & 1769   & 6405   & 0/60000       & 2/60000           \\ \hline
DDS 3 & 1382   & 1759   & 6274   & 0/60000       & 0/60000           \\ \hline
\end{tabular}
\end{table*}

\begin{table*}[ht]
\centering
\caption{Round-trip latency results: 32 KB of payload with RT settings}
\label{tab:32k_rt}
\begin{tabular}{|c|c|c|c|c|c|}
\hline
\multicolumn{6}{|c|}{32 KB of payload with RT settings} \\ \hline
      & Min($\mu$s) & Avg($\mu$s) & Max($\mu$s) & Missed deadlines & Message loss \\ \hline
DDS 1 & 2808    & 3226   & 3839   & 0/60001       & 0/60001           \\ \hline
DDS 2 & 4218  & 5505   & 6012   & 0/30000       & 0/30000           \\ \hline
DDS 3 & 10006   & 11655   & 12722   & 0/30001       & 0/30001           \\ \hline
\end{tabular}
\end{table*}

\begin{table*}[ht]
\centering
\caption{Round-trip latency results: System under load, 32 KB of payload and RT settings}
\label{tab:32k_stress_rt}
\begin{tabular}{|c|c|c|c|c|c|}
\hline
\multicolumn{6}{|c|}{System under load, 32 KB of payload and RT settings} \\ \hline
      & Min($\mu$s) & Avg($\mu$s) & Max($\mu$s) & Missed deadlines & Message loss \\ \hline
DDS 1 & 2913    & 3292   & 4961   & 0/60001       & 0/60001           \\ \hline
DDS 2 & 3991  & 4664   & 7175   & 0/30000       & 0/30000           \\ \hline
DDS 3 & 9828   & 10809   & 35285   & 3/30000       & 0/30000           \\ \hline
\end{tabular}
\end{table*}

\begin{table*}[ht]
\centering
\caption{Round-trip latency results: System under load, 128 KB of payload and RT settings}
\label{tab:128k_stress_rt}
\begin{tabular}{|c|c|c|c|c|c|}
\hline
\multicolumn{6}{|c|}{System under load, 128 KB of payload and RT settings} \\ \hline
      & Min($\mu$s) & Avg($\mu$s) & Max($\mu$s) & Missed deadlines & Message loss \\ \hline
DDS 1 & 9000    & 9369   & 179909   & 2/30000       & 75/30000           \\ \hline
DDS 2 & 11511  & 14151   & 176074   & 1991/18241       & 1192/18241           \\ \hline
DDS 3 & 35832   & 37238   & 48398   & 0/14997       & 0/14997           \\ \hline
\end{tabular}
\end{table*}

\begin{table*}[ht]
\centering
\caption{Round-trip latency results: System under load with 40 Mbps concurrent traffic and RT settings 12h}
\label{tab:128k_stress_rt}
\begin{tabular}{|c|c|c|c|c|c|}
\hline
\multicolumn{6}{|c|}{System under load with 40 Mbps concurrent traffic and RT settings 12h} \\ \hline
      & Min($\mu$s) & Avg($\mu$s) & Max($\mu$s) & Missed deadlines & Message loss \\ \hline
DDS 1 & 985    & 1283   & 4942   & 0/4319751       & 3/4319751           \\ \hline
DDS 2 & 1319  & 1808   & 6075   & 0/4320001       & 0/4320001           \\ \hline
DDS 3 & 1398   & 1803   & 6906   & 0/4320000       & 5/4320000           \\ \hline
\end{tabular}
\end{table*}

\begin{table*}[ht]
\centering
\caption{Round-trip latency results: System with 1 Mbps concurrent traffic and RT settings}
\label{tab:128k_stress_rt}
\begin{tabular}{|c|c|c|c|c|c|}
\hline
\multicolumn{6}{|c|}{System with 1 Mbps concurrent traffic and RT settings} \\ \hline
      & Min($\mu$s) & Avg($\mu$s) & Max($\mu$s) & Missed deadlines & Message loss \\ \hline
DDS 1 & 1048    & 1353   & 1587   & 0/60001       & 3/60001           \\ \hline
DDS 2 & 1418  & 1821   & 2274   & 0/60001       & 0/60001           \\ \hline
DDS 3 & 1336   & 1781   & 2089   & 0/60001       & 0/60001           \\ \hline
\end{tabular}
\end{table*}

\begin{table*}[ht]
\centering
\caption{Round-trip latency results: System under load with 1 Mbps concurrent traffic and RT settings}
\label{tab:128k_stress_rt}
\begin{tabular}{|c|c|c|c|c|c|}
\hline
\multicolumn{6}{|c|}{System under load with 1 Mbps concurrent traffic and RT settings} \\ \hline
      & Min($\mu$s) & Avg($\mu$s) & Max($\mu$s) & Missed deadlines & Message loss \\ \hline
DDS 1 & 1015    & 1310   & 1670   & 0/60001       & 0/60001           \\ \hline
DDS 2 & 1409  & 1726   & 2411   & 0/60001       & 0/60001           \\ \hline
DDS 3 & 1443   & 1788   & 2603   & 0/60001       & 0/60001           \\ \hline
\end{tabular}
\end{table*}


%





\end{document}